\documentclass{article}

\usepackage{arxiv}

\usepackage[utf8]{inputenc} 
\usepackage[T1]{fontenc}    
\usepackage{hyperref}       
\usepackage{url}            
\usepackage{booktabs}       
\usepackage{amsfonts}       
\usepackage{nicefrac}       
\usepackage{microtype}      
\usepackage{graphicx}
\usepackage[numbers]{natbib}
\usepackage{doi}
\usepackage{caption}
\usepackage{threeparttable}
\usepackage{placeins}
\usepackage{float} 
\usepackage[affil-it]{authblk} 
\usepackage{hyperref} 

\title{Biomedical Large Languages Models Seem not to be Superior to Generalist Models on Unseen Medical Data}

\author[1,2]{\textbf{Felix J. Dorfner} \textsuperscript{*}}
\author[3]{Amin Dada}
\author[4]{Felix Busch}
\author[5]{Tianyu Han}
\author[5]{Daniel Truhn}
\author[3,6,7,8]{Jens Kleesiek}
\author[9]{Madhumita Sushil}
\author[10]{Jacqueline Lammert}
\author[4]{Marcus R. Makowski}
\author[4]{\textbf{Lisa C. Adams}\textsuperscript{$\ddagger$}}
\author[4,11]{\textbf{Keno K. Bressem}\textsuperscript{$\dagger$}\textsuperscript{$\ddagger$}}

\affil[1]{Department of Radiology, Charité - Universitätsmedizin Berlin corporate member of Freie Universität Berlin and Humboldt Universität zu Berlin, Hindenburgdamm 30, 12203 Berlin, Germany}
\affil[2]{Athinoula A. Martinos Center for Biomedical Imaging, Massachusetts General Hospital and Harvard Medical School, 149 Thirteenth St, Charlestown, MA 02129, USA}
\affil[3]{Institute for AI in Medicine (IKIM), University Hospital Essen (AöR), Essen, Germany}
\affil[4]{Department of Diagnostic and Interventional Radiology, Technical University of Munich, School of Medicine and Health, Klinikum rechts der Isar, TUM University Hospital, Ismaninger Strasse 22, 81675, Munich, Germany}
\affil[5]{Department of Diagnostic and Interventional Radiology, University Hospital Aachen, Pauwelsstr. 30, 52074 Aachen, Germany}
\affil[6]{Cancer Research Center Cologne Essen (CCCE), West German Cancer Center Essen, University Hospital Essen (AöR), Essen, Germany}
\affil[7]{German Cancer Consortium (DKTK, Partner site Essen), Heidelberg, Germany}
\affil[8]{Department of Physics, TU Dortmund, Dortmund, Germany}
\affil[9]{Bakar Computational Health Sciences Institute, University of California, San Francisco, San Francisco}
\affil[10]{Department of Gynecology and Center for Hereditary Breast and Ovarian Cancer, Technical University of Munich, School of Medicine and Health, Klinikum rechts der Isar, TUM University Hospital, Ismaninger Strasse 22, 81675, Munich, Germany}
\affil[11]{Department of Cardiovascular Radiology and Nuclear Medicine, Technical University of Munich, School of Medicine and Health, German Heart Center, TUM University Hospital, Lazarethstr. 36, 80636, Munich, Germany}

\date{}

\renewcommand{\headeright}{}
\renewcommand{\shorttitle}{Biomedical Large Languages Models Seem not to be Superior to Generalist Models on Unseen Medical Data}

\hypersetup{
pdftitle={Biomedical Large Languages Models Seem not to be Superior to Generalist Models on Unseen Medical Data},
pdfsubject={Large Language Models, Radiology, Natural Language Processing},
pdfauthor={Felix J. Dorfner, Amin Dada, Felix Busch, Marcus R. Makowski, Tianyu Han, Daniel Truhn, Jens Kleesiek, Madhumita Sushil, Jacqueline Lammert, Lisa C. Adams, Keno K. Bressem},
pdfkeywords={Large Language Models, Radiology, Natural Language Processing},
}

\begin{document}
\maketitle
\textsuperscript{*}First author.\\
\textsuperscript{$\dagger$}Corresponding author: keno.bressem@tum.de \\
\textsuperscript{$\ddagger$}These authors contributed equally as last authors.

\abstract
Large language models (LLMs) have shown potential in biomedical applications, leading to efforts to fine-tune them on domain-specific data. However, the effectiveness of this approach remains unclear. This study evaluates the performance of biomedically fine-tuned LLMs against their general-purpose counterparts on a variety of clinical tasks. We evaluated their performance on clinical case challenges from the New England Journal of Medicine (NEJM) and the Journal of the American Medical Association (JAMA) and on several clinical tasks (e.g., information extraction, document summarization, and clinical coding). Using benchmarks specifically chosen to be likely outside the fine-tuning datasets of biomedical models, we found that biomedical LLMs mostly perform inferior to their general-purpose counterparts, especially on tasks not focused on medical knowledge. While larger models showed similar performance on case tasks (e.g., OpenBioLLM-70B: 66.4\% vs. Llama-3-70B-Instruct: 65\% on JAMA cases), smaller biomedical models showed more pronounced underperformance (e.g., OpenBioLLM-8B: 30\% vs. Llama-3-8B-Instruct: 64.3\% on NEJM cases). Similar trends were observed across the CLUE (Clinical Language Understanding Evaluation) benchmark tasks, with general-purpose models often performing better on text generation, question answering, and coding tasks. Our results suggest that fine-tuning LLMs to biomedical data may not provide the expected benefits and may potentially lead to reduced performance, challenging prevailing assumptions about domain-specific adaptation of LLMs and highlighting the need for more rigorous evaluation frameworks in healthcare AI. Alternative approaches, such as retrieval-augmented generation, may be more effective in enhancing the biomedical capabilities of LLMs without compromising their general knowledge.

\endabstract

\section{Introduction}\label{sec1}

Large language models (LLMs) have shown remarkable potential for various applications, including in the biomedical domain.\cite{RN1,RN2,RN3} These models can serve as knowledge sources, aid in information retrieval from patient notes, assist with data structuring or coding, and support patient anamnesis. \cite{RN3} To enhance LLMs' performance on domain-specific tasks, several initiatives have focused on fine-tuning these models using biomedical data.\cite{RN4,RN5,RN6,RN7,RN8,RN9}
Recent approaches include BioMistral-7b,\cite{RN6} based on Mistral 7b,\cite{RN10} and OpenBioLLM, \cite{RN5}  which uses the Llama3 (Large Language Model Meta AI) models as a foundation.\cite{RN11} These efforts aim to create specialized models for biomedical applications.
However, assessing the performance of fine-tuned models presents challenges. While using exam questions, such as those from the United States Medical Licensing Examination (USMLE), is common, \cite{RN5,RN6,RN7,RN9,RN12} this approach may not accurately reflect a model's performance in real clinical practice. Additionally, the prolonged availability of these questions online risks data leakage and benchmark corruption.\cite{RN13}
The latest general-purpose LLMs, like Llama 3 and Mistral, are trained on vast amounts of web data, likely including substantial medical information. This raises questions about the added value of fine-tuning, with limited availability of domain-specific data, research groups may struggle to introduce novel information not already present in the training data of large AI companies. Consequently, biomedical LLMs fine-tuned on similar web-based data might face issues such as redundant learning or even performance degradation due to catastrophic forgetting. \cite{RN14}
To address these concerns, we conducted a comparative study of recent biomedical LLMs and their general-purpose baseline models. Our evaluation used data from recent benchmarks, specifically chosen to likely be outside the fine-tuning process of the biomedical models, ensuring a fair assessment. Given the domain-specific nature of this data, we hypothesized that the biomedical models outperform their general-purpose counterparts.
This study aims to provide insights into the effectiveness of domain-specific fine-tuning for biomedical LLMs and to contribute to the ongoing discussion about optimal strategies for adapting LLMs to specialized fields.

\section{Methods}\label{sec2}

The benchmarks used in this study were carefully selected to represent a wide range of clinical tasks while ensuring they were likely outside the fine-tuning datasets of biomedical models. We chose recently published case vignettes and newly developed benchmarks to minimize the risk of data contamination. The biomedical and general-purpose models were selected to represent the state-of-the-art in both categories, covering different model sizes and architectures to ensure a comprehensive comparison.

\subsection{Benchmarks}

\subsubsection{Clinical Case Challenges}
We evaluated the performance of biomedical and general-purpose LLMs using clinical challenges from two medical journals: the New England Journal of Medicine (NEJM) and the Journal of the American Medical Association (JAMA). These case vignettes represent real-world clinical scenarios and cover a wide range of medical specialties and conditions. The NEJM dataset consists of 347 questions, while the JAMA dataset contains 140 questions. These case vignettes, presented with multiple-choice answers, cover a wide range of medical knowledge and provide a comprehensive assessment of the models' performance on unseen medical data. Models are evaluated by accuracy of correctly answered questions.

\subsubsection{MeDiSumQA}
This task involves answering questions based on discharge summaries from the MIMIC-IV database.16 It tests the model's ability to extract relevant information from lengthy clinical documents and provide accurate, patient-friendly responses. \cite{RN15}

\subsubsection{MeDiSumCode}
This task evaluates the model's capability to assign appropriate ICD-10 codes to diagnoses and procedures mentioned in discharge summaries. The task was introduced together with MeDiSumQA as part of the CLUE benchmark and requires both accurate information extraction and a deep understanding of medical coding systems. \cite{RN15}

\subsubsection{MedNLI}
Based on the MIMIC-III dataset,\cite{RN16} this natural language inference task assesses the model's ability to determine the logical relationship between a premise (a sentence from a clinical note) and a hypothesis.\cite{RN17}

\subsubsection{MeQSum}
This task involves summarizing consumer health queries, testing the model's ability to understand lay language and reformulate it into concise, medically sound queries.\cite{RN18}

\subsubsection{ProblemSummary}
Using clinical notes organized according to the SOAP (Subjective, Objective, Assessment, Plan) principle, this task requires models to predict a patient's current health problems based on the Subjective and Assessment sections.\cite{RN19}

\subsubsection{LongHealth}
This task uses 20 fictional patient records to test the model's ability to handle long documents, answer questions about them, and identify when information is not available in the given context.\cite{RN20} Evaluation is split into three sub-tasks: a) Answering questions about long document. b) Handling increased input length with unrelated documents c) Identifying when information is not available.

MeDiSumQA, MeDiSumCode and LongHealth require model with longer context size, limiting the number of models that can be evaluated on these benchmarks. 
For text generation tasks (MeDiSumQA, MeQSum, and ProblemSummary), ROUGE scores (ROUGE-1, ROUGE-2, ROUGE-L) and BERTScore are used to evaluate the quality and semantic similarity of generated outputs. The MeDiSumCode task is assessed using F1-scores for exact and approximate matches of ICD-10 codes, as well as the ratio of valid codes generated. MedNLI uses accuracy to evaluate the model's ability to classify relationships between premises and hypotheses. For tasks involving entity recognition and medical concepts (ProblemSummary), F1-scores for UMLS entity extraction are also employed. The LongHealth task primarily uses accuracy across its subtasks to evaluate comprehension and question-answering abilities on long clinical documents. \cite{RN15}

\subsection{Models Evaluated}

\subsubsection{Generalist Models}
Llama (Large Language Models by Meta AI) is a series of foundation models developed by Meta AI. We evaluated models based on three generations of Llama. Llama 1, released in 2023, Llama 1 was trained on trillions of tokens from publicly available datasets.\cite{RN21} As the base model is no longer publicly accessible, Llama 1 was not evaluated directly. Llama 2, an improved version released later in 2023, Llama 2 features enhanced training on a larger dataset and improvements in model architecture.\cite{RN22} Llama 2 models were released as base models as well as finetuned chat models. Llama 3 Is the latest iteration in the Llama series, released in 2024.\cite{RN11} Llama 3 incorporates advanced training techniques and the largest training dataset of all Llama models. It was again released as base model and finetuned chat model. 
Mistral 7B is an open-source language model developed by Mistral AI, released in 2023, it utilizes novel architectural choices such as grouped-query attention and sliding window attention, allowing for efficient processing of long sequences.\cite{RN10} For all base models, the  chat/instruction tuned versions, provided by the developers were used (Llama-2-7b-chat-hf, Llama-2-70b-chat-hf, Llama-3-8B-Instruct, Llama-3-70B-Instruct and Mistral-7B-Instruct-v0.2).

\subsubsection{Biomedical models based on Llama}
PMC-Llama-7B, based on Llama 1, was fine-tuned on biomedical literature from PubMed Central.7 MedAlpaca-7B, also Llama 1-based, was  fine-tuned on a curated dataset of medical information, including medical wikis and medical flash cards.\cite{RN9}  Meditron-7B, med42 and ClinicalCamel-70B are models based on Llama 2 with 7B parameters or 70B parameters.\cite{RN8, RN23, RN24} OpenBioLLM-70B and OpenBioLLM-8B are the most recent biomedical LLMs, based on Llama 3. They currently achieve state of the art on USMLE question answering. BioMistral-7B and JSL-MedMNX-7B-SFT are finetuned versions of Mistral-7B using biomedical data.\cite{RN6}

\subsection{Evaluation Procedure}

The evaluation of the models was conducted using a combination of methods tailored to each benchmark. For the clinical case challenges from NEJM and JAMA, we employed a GPU server equipped with four NVIDIA A100 GPUs. All models were evaluated in their version that is available through the Hugging Face transformers library (version 4.40.1). For compatible models, vLLM (version 0.4.2) was used, which provides a parallelized engine around the Hugging Face models for faster inference. All models were tested with identical parameters: temperature set to 0, top p at 0.95, and both frequency and presence penalties at 0. The 70B models were run across all four GPUs using Ray (version 2.21.0), while the smaller models were executed on a single A100 GPU. The models Meditron-7b MedAlpaca 7b, and PMC-Llama-7B did not produce comprehensible outputs in the initial prompt setup. As these models were trained with a fixed system prompt, the order of the prompt parts was changed to provide the actual clinical vignette at the very end. This was deemed necessary to ensure a fair evaluation of the model capabilities. Additionally, for PMC-Llama-7B the developer recommended settings using top-k sampling with k=50 had to be used to obtain valid outputs.
All other benchmark evaluations were performed on an NVIDIA DGX node with 8 A100 GPUs. The Hugging Face Text Generation Inference Toolkit (v2.1.0) was used for model inference. Larger models were distributed across up to four GPUs, while smaller models were loaded onto a single GPU. Whenever the tokenizer provided a chat template, it was used to format the model input. For models where the template was described in the model card, the template was added manually. All models were tested with the Hugging Face default parameters (temperature 1, top p 0.95, and no penalties).

\section{Results}\label{sec3}

\subsection{Clinical Case Challenges}
On the JAMA case challenges, OpenBioLLM-70B achieved the highest accuracy at 66\%, followed by Llama-3-70B-Instruct at 65\%. JSL-MedMNX-7B-SFT achieved 54\% accuracy, slightly higher than its base model Mistral-7B-Instruct-v0.2 with 52\%. Notably, Llama-3-8B-Instruct (57\%) outperformed its biomedical counterpart OpenBioLLM-8B (18\%). Also, the Llama 1 based model MedAlpaca 7b outperformed OpenBioLLM-8B with 22\% accuracy. BioMistral-7B (28\%) underperformed compared to Mistral-7B-Instruct-v0.2. In contrast med42-70B and JSL-MedMNX-7B-SFT performed better than their respective baseline models.  
For the NEJM cases, OpenBioLLM-70B and Llama-3-70B-Instruct both achieved 74\% accuracy. Llama-3-8B-Instruct (64\%) again outperformed OpenBioLLM-8B (30\%). JSL-MedMNX-7B-SFT and Mistral-7B-Instruct-v0.2 performed similarly (47\% and 46\% respectively). BioMistral-7B (37\%) again underperformed compared to its base model. PMC-Llama-7B and Meditron-7b showed poor performance on both datasets. Table \ref{tab:table_1} provides an overview of model accuracy in the clinical case vignettes. Figure \ref{fig:figure1} provide an overview of the performance on all evaluated tasks. 

\begin{table}[!ht]
    \centering
    \caption{Individual results on the clinical case challenges.}
    \begin{tabular}{lll}
        \specialrule{1pt}{0pt}{0pt}

        ~ & JAMA Case Challenges & NEJM Case Challenges \\ \hline
        \textbf{Llama 1 Models} & ~ & ~ \\ \hline
        MedAlpaca 7B & \textbf{22.1\% (15\%-29\%)} & \textbf{17.3\% (13\%-21\%)} \\ 
        PMC-Llama-7B & 4.3\% (1\%-8\%) & 2.3\% (1\%-4\%) \\ \hline
        \textbf{Llama 2 Models} & ~ & ~ \\ \hline
        Llama-2-7b-chat-hf & 44.3\% (36\%-53\%) & 26.8\% (22\%-32\%) \\ 
        Llama-2-70b-chat-hf & 45.7\% (38\%-54\%) & 43.5\% (38\%-49\%) \\
        Meditron-7B & 12.9\% (7\%-18\%) & 4.9\% (3\%-7\%) \\ 
        ClinicalCamel-70b & 49.3\% (41\%-57\%) & 38.9\% (34\%-44\%) \\ 
        Med42-70b & \textbf{52.9\% (44\%-61\%)} & \textbf{56.5\% (51\%-61\%)} \\ \hline
        \textbf{Llama 3 Models} & ~ & ~ \\ \hline
        Llama-3-8B-Instruct & 57.1\% (49\%-65\%) & 64.3\% (59\%-69\%) \\ 
        Llama-3-70B-Instruct & 65\% (57\%-73\%) & \textbf{74.6\% (70\%-79\%)} \\ 
        OpenBioLLM-8B & 17.9\% (12\%-24\%) & 30\% (25\%-35\%) \\ 
        OpenBioLLM-70B & \textbf{66.4\% (59\%-74\%)} & 74.1\% (70\%-78\%) \\ \hline
        \textbf{Mistral -7B Models} & ~ & ~ \\ \hline
        Mistral-7B-Instruct-v0.2 & 52.1\% (44\%-60\%) & 46.4\% (41\%-52\%) \\ 
        JSL-MedMNX-7B-SFT & \textbf{53.6\% (45\%-63\%)} & \textbf{46.7\% (41\%-52\%)} \\ 
        Biomistral-7B & 27.9\% (21\%-35\%) & 37.2\% (32\%-42\%) \\ 
        \specialrule{1pt}{0pt}{0pt}

    \end{tabular}
    \label{tab:table_1}
\end{table}

\subsection{MedNLI Task}
OpenBioLLM-70B achieved the highest accuracy with 80.85\%, closely followed by the generalist model Llama-3-70B-Instruct (79.37\%) and JSL-MedMNX-7b-SFT (79.3\%). Notably, OpenBioLLM-8B (44.93\%) underperformed compared to Llama-3-8B-Instruct (74.08\%). BioMistral-7B (62.75\%) also showed lower accuracy than its base model Mistral-7B-Instruct-v0.2 (69.93\%).

\subsection{ProblemSummary}
Llama-2-7B-chat-hf achieved the highest BERT F1 score (75.98) on MeQSum, slightly outperforming larger and more recent models. All biomedical LLMs underperformed compared to their respective generalist models across all metrics.

\subsection{LongHealth}
Models based on Llama 1 and Llama 2 did not support enough context, to be evaluated on LongHealth, MeDiSumQA and MeDiSumCode. Llama-3-70B-Instruct demonstrated superior performance across all LongHealth tasks. OpenBioLLM-70B showed lower scores than its base model, particularly in Task 3, which evaluates how prone to hallucinating non-existing information models are. JSL-MedMNX-7b-SFT underperformed compared to Mistral-7B-Instruct-v0.2, except for Task 3, where JSL-MedMNX-7b-SFT achieved a higher score than Mistral-7B-Instruct-v0.2. BioMistral-7B showed consistently worse performance compared to Mistral-7B-Instruct-v0.2. 

\subsection{MeDiSumQA}
Llama-3-70B-Instruct achieved the highest scores across all MeDiSumQA metrics. OpenBioLLM-70B showed comparable but slightly lower performance. BioMistral-7B and JSL-MedMNX-7b-SFT both underperformed compared to Mistral-7B-Instruct-v0.2, with notably lower ROUGE and BERT F1 scores.

\subsection{MeDiSumQA}
Llama-3-70B-Instruct significantly outperformed all other models in MeDiSumCode, achieving the highest scores across all metrics. OpenBioLLM-70B showed lower performance than its base model, particularly in EM F1 and AP F1 scores. BioMistral-7B underperformed compared to Mistral-7B-Instruct-v0.2 (68.76), while JSL-MedMNX-7b-SFT (68.48) showed similar performance to its base model in Valid Code Accuracy.

\begin{figure}[ht]
    \centering
    \includegraphics[width=0.85\linewidth]{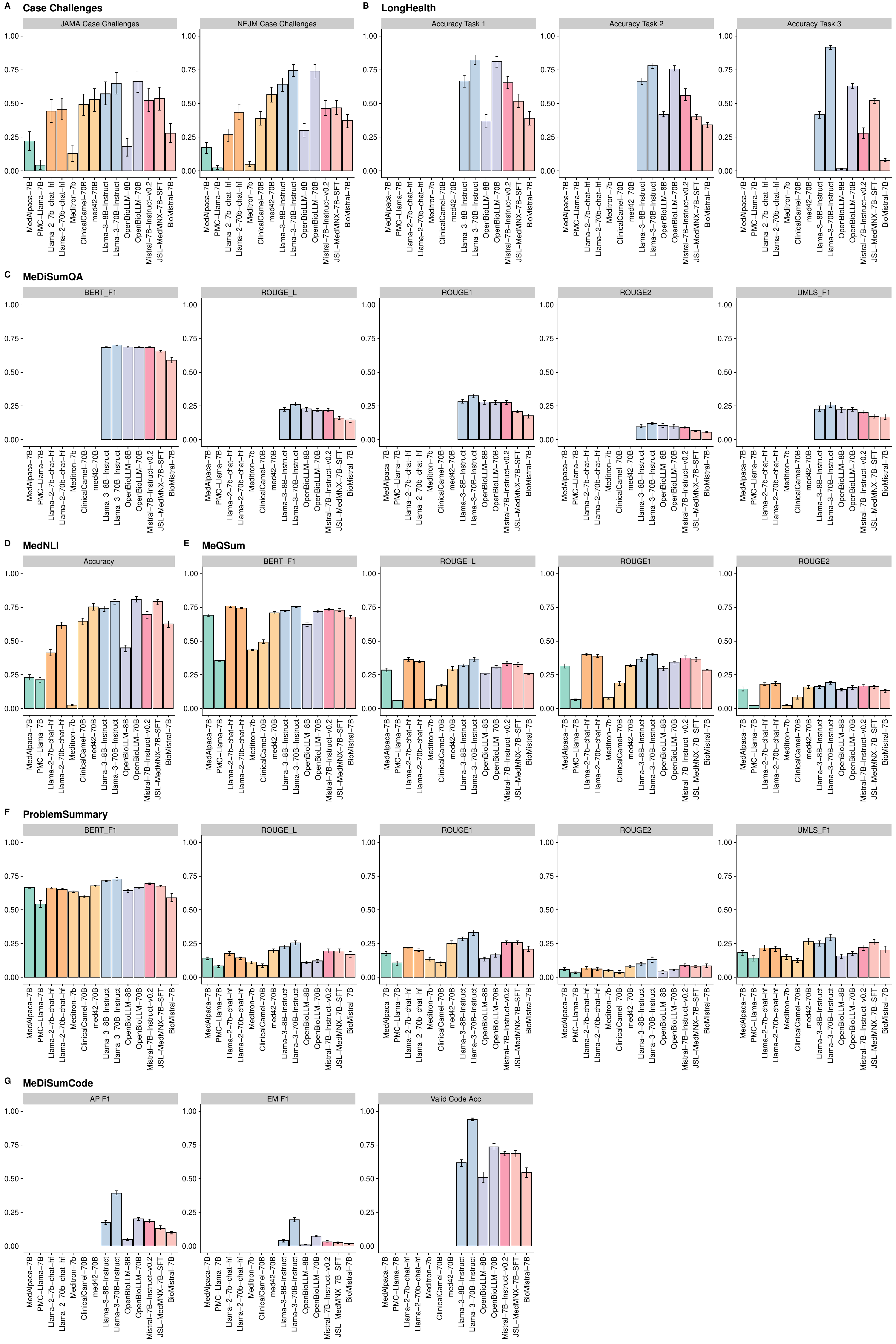}
    \caption{Comparative performance of the domain specific and general-purpose base models on the two clinical case vignette series. Bars represent the respective metric and error bars represent 95\% confidence intervals obtained through bootstrapping.}
    \label{fig:figure1}
\end{figure}

Tables \ref{tab:table_2} and \ref{tab:table_3} show individual metrics of all tasks except clinical case vignettes. Figure \ref{fig:figure1} provides an overview of all tasks and models.  

\begin{table}[!ht]
    \centering
    \caption{Results on MedNLI, ProblemSummary and MeQSum.}
    \resizebox{0.95\textwidth}{!}{
    \begin{tabular}{lc|c|cccc|cccccc}
        \specialrule{1pt}{0pt}{0pt}
        ~ &  \multicolumn{1}{c}{\textbf{}} & \multicolumn{1}{c}{\textbf{MedNLI}} & \multicolumn{4}{c}{\textbf{ProblemSummary}} & \multicolumn{5}{c}{\textbf{MeQSum}} \\ \hline
        ~ & Mean score & Acc & R-L & R-1 & R-2 & BERT F1 & UMLS F1 & R-L & R-1 & R-2 & BERT F1 \\ \hline
        \textbf{Llama 1 Models} & ~ & ~ & ~ & ~ & ~ & ~ & ~ & ~ & ~ & ~ & ~ \\ \hline
        MedAlpaca 7B & \textbf{27.24} & \textbf{22.89} & \textbf{14.15} & \textbf{17.59} & \textbf{5.67} & \textbf{66.43} & \textbf{18.32} & \textbf{28.43} & \textbf{31.48} & \textbf{14.51} & \textbf{69.21} \\ 
        PMC-Llama-7B & 17.30 & 21.2 & 8.35 & 10.57 & 3.36 & 54.39 & 14.21 & 6.03 & 6.64 & 2.11 & 35.35 \\ \hline
        \textbf{Llama 2 Models} & ~ & ~ & ~ & ~ & ~ & ~ & ~ & ~ & ~ & ~ & ~ \\ \hline
        Llama-2-7b-chat-hf & 36.94 & 41.27 & \textbf{17.43} & \textbf{22.25} & \textbf{6.93} & \textbf{66.33} & \textbf{21.62} & \textbf{36.45} & \textbf{39.99} & 18.13 & \textbf{75.98} \\ 
        Llama-2-70b-chat-hf & \textbf{42.90} & 61.69 & 14.43 & 19.81 & 6.14 & 65.07 & 21.37 & 34.91 & 38.81 & \textbf{18.48} & 74.37 \\ 
        Meditron-7B & 13.03 & 2.39 & 11.29 & 13.43 & 4.81 & 63.35 & 15.21 & 6.83 & 7.90 & 2.26 & 43.39 \\ 
        ClinicalCamel-70b & 35.71 & \textbf{64.65} & 8.62 & 10.83 & 3.71 & 59.94 & 12.34 & 16.96 & 18.8 & 8.49 & 49.3 \\ \hline
        \textbf{Llama 3 Models} & ~ & ~ & ~ & ~ & ~ & ~ & ~ & ~ & ~ & ~ & ~ \\ \hline
        Llama-3-8B-Instruct & 48.37 & 74.08 & 22.7 & 28.52 & 9.87 & 71.45 & 25.32 & 32.2 & 36.49 & 16.37 & 72.74 \\ \hline
        Llama-3-70B-Instruct & \textbf{52.36} & 79.37 & \textbf{25.43} & \textbf{33.16} & \textbf{13.01} & \textbf{73.00} & \textbf{29.12} & \textbf{36.57} & \textbf{40.20} & \textbf{19.30} & \textbf{75.74} \\ 
        OpenBioLLM-8B & 33.20 & 44.93 & 10.82 & 13.62 & 4.03 & 64.14 & 15.67 & 26.21 & 29.41 & 14.03 & 62.39 \\ 
        OpenBioLLM-70B & 47.57 & \textbf{80.85} & 12.10 & 16.67 & 5.58 & 66.51 & 17.74 & 30.72 & 34.31 & 15.55 & 71.99 \\ 
        Med 42 & 47.29 & 75.42 & 19.73 & 25.14 & 7.75 & 67.84 & 26.36 & 29.34 & 32.07 & 15.96 & 70.96 \\ \hline
        \textbf{Mistral-7B Models} & ~ & ~ & ~ & ~ & ~ & ~ & ~ & ~ & ~ & ~ & ~ \\ \hline
        Mistral-7B-Instruct-v0.2 & 46.45 & 69.93 & \textbf{19.57} & \textbf{25.59} & \textbf{8.92} & \textbf{69.64} & 22.07 & \textbf{33.54} & \textbf{37.47} & \textbf{16.61} & \textbf{73.47} \\ 
        JSL-MedMNX-7B-SFT & \textbf{49.40} & \textbf{79.3} & 19.49 & 25.57 & 8.21 & 67.62 & \textbf{25.72} & 32.49 & 36.58 & 16.25 & 73.01 \\ 
        Biomistral-7B & 40.57 & 62.75 & 16.90 & 20.89 & 8.4 & 59.03 & 20.12 & 25.89 & 28.46 & 13.31 & 67.93 \\ 
        \specialrule{1pt}{0pt}{0pt}

    \end{tabular}
    }
    \label{tab:table_2}
\end{table}

\begin{table}[!ht]
    \centering
    \caption{Results on MedNLI, ProblemSummary and MeQSum.}
    \resizebox{0.95\textwidth}{!}{
    \begin{tabular}{lc|ccc|ccccc|ccc}
        \specialrule{1pt}{0pt}{0pt}
        ~ & \multicolumn{1}{c}{} & \multicolumn{3}{c}{\textbf{LongHealth}} & \multicolumn{5}{c}{\textbf{MeDiSumQA}} & \multicolumn{3}{c}{\textbf{MeDiSumCode}} \\ \hline
        ~ & Mean score & Task 1 & Task 2 & Task 3 & R-L & R-1 & R-2 & BERT F1 & UMLS F1 & EM F1 & AP F1 & Valid Code Acc \\ \hline
        \textbf{Llama 3 Models} & ~ & ~ & ~ & ~ & ~ & ~ & ~ & ~ & ~ & ~ & ~ & ~ \\ \hline
        Llama-3-8B-Instruct & 40.61 & 68.3 & 66.55 & 56.25 & 22.44 & 28.15 & 9.63 & 68.62 & 22.74 & 3.95 & 17.55 & 61.93 \\ 
        Llama-3-70B-Instruct & \textbf{56.00} & \textbf{81.65} & \textbf{77.90} & \textbf{91.70 }& \textbf{26.20} & \textbf{32.5} & \textbf{11.93} & \textbf{70.24} & \textbf{25.78} & \textbf{19.65} & \textbf{39.20} & \textbf{93.94}\\ 
        OpenBioLLM-8B & 25.44 & 37.55 & 41.75 & 1.55 & 22.89 & 27.95 & 10.45 & 68.7 & 22.15 & 0.84 & 4.84 & 51.16 \\ 
        OpenBioLLM-70B & 45.56 & 80.2 & 75.60 & 62.90 & 21.85 & 27.8 & 9.54 & 68.43 & 22.47 & 7.37 & 20.24 & 73.65 \\ 
        \textbf{Mistral-7B Models} & ~ & ~ & ~ & ~ & ~ & ~ & ~ & ~ & ~ & ~ & ~ & ~ \\ \hline
        Mistral-7B-Instruct-v0.2 & \textbf{38.94} & \textbf{67.2 }& \textbf{62.4}& 42.45 & \textbf{21.8} & \textbf{27.47} & \textbf{9.19} & \textbf{68.43} & \textbf{20.30} & \textbf{3.08} & \textbf{18.23} & \textbf{68.76} \\ 
        JSL-MedMNX-7B-SFT & 34.03 & 53.15 & 40.2 & \textbf{52.25} & 15.93 & 20.89 & 6.7 & 65.9 & 17.33 & 2.82 & 13.3 & 68.48 \\ 
        Biomistral-7B & 23.83 & 38.05 & 34.25 & 7.8 & 14.65 & 17.81 & 5.46 & 59.01 & 16.88 & 1.67 & 9.92 & 54.51 \\ 
        \specialrule{1pt}{0pt}{0pt}
    \end{tabular}
    }
    \label{tab:table_3}
\end{table}

\section{Discussion}\label{sec4}
Our comprehensive evaluation of LLMs across clinical tasks has yielded unexpected insights regarding the performance of biomedically fine-tuned models. Contrary to our initial hypothesis, biomedical models generally underperformed compared to their general-purpose counterparts across various tasks. This suggests that fine-tuning LLMs on biomedical data may not provide the expected benefits and may even decrease rather than improve performance.

Several factors may contribute to this underperformance. The superior performance of general-purpose models might stem from their exposure to a more diverse range of topics and linguistic structures during pre-training. This broader knowledge base could enable more flexible reasoning and better generalization to novel tasks. Additionally, the fine-tuning process for biomedical models might inadvertently introduce biases or overly narrow the models’ focus, potentially limiting their ability to integrate broader contextual information crucial for complex clinical reasoning.
Another possibility is overfitting to specific medical datasets used during fine-tuning, leading to reduced generalization capabilities. This could be due to data leakage between training and test sets, \cite{RN25} a risk that increases with the growing size of training datasets, making it increasingly challenging for researchers to verify data integrity, especially as some models evaluated (OpenBioLLM or JSL-MedMNX-7b-SFT) do not report their training data. Overfitting can also occur indirectly through repeated evaluation on common test datasets, such as USMLE or MMLU, which may inadvertently select for models that perform well on these specific benchmarks.
A key factor to consider is the potential loss of general knowledge during the fine-tuning process, a phenomenon known as catastrophic forgetting.\cite{RN14} Our findings highlight the delicate balance required when adapting general-purpose models to specialized domains without compromising their broad capabilities. Potential evidence supporting this hypothesis comes from the observation that biomedical LLMs that only underwent supervised fine-tuning (medAlpaca-7B, ClinicalCamel-70B, OpenBioLLM-8B/70B)\cite{RN5, RN9, RN23} showed a smaller performance decrease compared to their general-purpose counterparts than models that underwent continued pretraining (BioMistral-7B, Meditron-7B). \cite{RN6, RN23}

Interestingly, larger models exhibited smaller performance gaps between biomedical and generalist versions. Given that both were trained with the same amount of data, the overall changes in model weights of larger models might have been smaller, potentially reducing the risk of catastrophic forgetting. These observations suggest that using only fine-tuning, rather than continued pretraining, may be preferable when adapting LLMs for specific domains. However, the data presented in our analysis is insufficient for definitive conclusions, and further research is needed to investigate these phenomena thoroughly.

The consistent strong performance of Llama-3-70B-Instruct across all benchmarks is particularly noteworthy. As the latest iteration of Llama models, Llama 3 was trained on an unprecedented 15 trillion tokens of data, a sevenfold increase compared to Llama 2 and substantially more than previously believed optimal.\cite{RN11,RN22,RN26} With this vast amount of data, it is highly likely that nearly all freely available biomedical texts on the internet are included in the training data of these general-purpose models, enabling them to inherently capture sufficient medical knowledge. Consequently, unless biomedical LLMs are fine-tuned on novel, previously unavailable data (e.g., copyrighted scientific papers, proprietary hospital data), fine-tuning on publicly accessible biomedical data may not add new knowledge. Instead, it may risk the model forgetting valuable information through continued fine-tuning. 
Performance differences between biomedical and general LLMs vary depending on the task. While the performance of OpenBioLLM 70B and Llama3 70B appears to be on par for the clinical vignettes, which have a uniform multiple-choice format, OpenBioLLM performed significantly worse than Llama3 70B on the MeDiSumCode benchmark, which requires in-depth knowledge of the ICD coding system with over 70,000 individual codes. This discrepancy suggests that the benefits of biomedical fine-tuning may be task-dependent, with potentially greater advantages in highly specialized medical tasks that require deep domain knowledge.

One critical finding of our study is the higher risk of hallucinations observed in biomedical LLMs. The results from LongHealth Task 3, which evaluates hallucination tendencies, raise important concerns about the reliability of LLMs in clinical applications. The superior performance of general-purpose models in this aspect is particularly intriguing and warrants further investigation. Recent work has highlighted the critical nature of this issue in healthcare AI, emphasizing the need for robust strategies to mitigate hallucination risks.\cite{RN27,RN28}

These results could have significant implications for the development and application of LLMs in healthcare. They challenge the prevailing assumption that domain-specific fine-tuning is universally beneficial for specialized tasks. Instead, our findings suggest that the relationship between model performance and domain adaptation is more nuanced and complex than previously thought. This complexity may stem from the intricate interplay between a model's general knowledge and its ability to apply that knowledge in specific contexts.

Interestingly, larger models exhibited smaller performance gaps between biomedical and generalist versions. Given that both were trained with the same amount of data, the overall changes in model weights of larger models might have been smaller, potentially reducing the risk of catastrophic forgetting. These observations suggest that using only fine-tuning, rather than continued pretraining, may be preferable when adapting LLMs for specific domains. However, further research is needed to investigate these observations more thoroughly.

\subsection*{Limitations }
Our comparison has limitations. The two sets of case vignettes have been freely available on the web and might thus have partly been included in the training data of recent LLMs, leading to an overestimation of the LLM performance. However, since the domain-specific models are based on the general-purpose LLMs, this would not be an advantage. Rather, the inferior performance of the biomedical LLMs, could be an additional argument for the presence of catastrophic forgetting. Furthermore, while the benchmarks used cover a range of clinical tasks, they may not fully represent the complexity and diversity of real-world clinical scenarios. Specifically, they do not cover detailed medical knowledge such as nuanced diagnostic criteria, extensive patient history considerations, and comprehensive treatment recommendations.

\subsection*{Conclusion}
In conclusion, our study challenges prevailing assumptions about the effectiveness of biomedical fine-tuning for LLMs, with potential implications for domain-specific adaptation in general. Rather than continued pre-training or fine-tuning, alternative approaches such as retrieval-augmented generation are worth exploring to enhance the biomedical capabilities of LLMs without compromising their general knowledge. Recent studies have shown promising results for these techniques.\cite{RN29,RN30} 
While biomedical LLMs perform well on widely used benchmarks such as the USMLE or MMLU, our evaluation reveals variable performance across tasks and models. This highlights the need for more rigorous, task-specific evaluation frameworks for healthcare LLMs. These evaluations should focus on clinical support tasks such as text summarization, information retrieval, and data structuring. We believe that using LLMs for these applications could provide more noticeable relief to healthcare workers than using LLMs as (potentially unreliable) knowledge bases.

\FloatBarrier

\newpage
\bibliographystyle{unsrtnat}
\bibliography{references.bib}  

\newpage

\end{document}